\RequirePackage{fix-cm}
\documentclass{svjour3}                     
\smartqed  
\usepackage{amsmath,amssymb,amsfonts,bm}
\usepackage{graphicx}
\usepackage{multirow}
\usepackage[misc]{ifsym}
\usepackage{booktabs}
\usepackage{subfigure}
\usepackage{natbib}
\usepackage{color}
\usepackage[colorlinks,linkcolor=blue,anchorcolor=blue,citecolor=blue,urlcolor=blue]{hyperref}
\usepackage{geometry}
\begin{document}

\title{A physics and data co-driven surrogate modeling approach for temperature field prediction on irregular geometric domain
}


\author{Kairui Bao \textsuperscript{1}   \and
        Wen Yao \textsuperscript{2,*} \and 
        Xiaoya Zhang \textsuperscript{2,*} \and 
        Wei Peng \textsuperscript{2} \and
        Yu Li \textsuperscript{2}
}


\institute{
	\Letter{Wen Yao} \\
	\email{wendy0782@126.com} \\
	\at
	\Letter{Xiaoya Zhang} \\
	\email{zhangxiaoya09@nudt.edu.cn} \\
	\at
	\textsuperscript{1} College of Aerospace Science and Engineering, National University of Defense Technology, No. 109, Deya Road, Changsha 410073, China
	\at	
	\textsuperscript{2}Defense Innovation Institute, Chinese Academy of Military Science, Beijing 100071, China
}

\date{Received: date / Accepted: date}

\maketitle

\begin{abstract}

In the whole aircraft structural optimization loop, thermal analysis plays a very important role. But it faces a severe computational burden when directly applying traditional numerical analysis tools, especially when each optimization involves repetitive parameter modification and thermal analysis followed. Recently, with the fast development of deep learning, several Convolutional Neural Network (CNN) surrogate models have been introduced to overcome this obstacle. 
However, for temperature field prediction on irregular geometric domains (TFP-IGD), CNN can hardly be competent since most of them stem from processing for regular images. 
To alleviate this difficulty, we propose a novel physics and data co-driven surrogate modeling method. First, after adapting the Bezier curve in geometric parameterization, a body-fitted coordinate mapping is introduced to generate coordinate transforms between the irregular physical plane and regular computational plane. 
Second, a physics-driven CNN surrogate with partial differential equation (PDE) residuals as a loss function is utilized for fast meshing (meshing surrogate); then, we present a data-driven surrogate model based on the multi-level reduced-order method, aiming to learn solutions of temperature field in the above regular computational plane (thermal surrogate). 
Finally, combining the grid position information provided by the meshing surrogate with the scalar temperature field information provided by the thermal surrogate (combined model), we reach an end-to-end surrogate model from geometric parameters to temperature field prediction on an irregular geometric domain. 
Numerical results demonstrate that our method can significantly improve accuracy prediction on a smaller dataset while reducing the training time when compared with other CNN methods.
\end{abstract}

\section{Introduction}
\label{intro}
Thermal analysis plays a vital role in the aircraft structure design and optimization \cite{bodie2010thermal,sanchez2020thermal,munk2017effect}. In the design process, each shape modification involves repetitive thermal analysis. Thus, a fast and accurate temperature field prediction could greatly help to improve the efficiency of aircraft design \cite{Yao2011,Zheng2020,Zheng2019}. An ideal circumstance is to reach real-time prediction while guaranteeing accuracy simultaneously. However, for most cases setting on the irregular geometric domain, it is hard to realize this due to the lack of analytical solutions, and many efforts have been made towards this end.

Since temperature field prediction heavily relies on solving partial differential equations (PDEs), traditional numerical methods, such as finite difference (FD), finite element (FE), and finite volume (FV) methods, are always employed. 
These numerical methods usually follow the same formula: first grid generation and then temperature fields solving. 
Most of the present commercial simulation software take this recipe, like ANSYS, ABAQUS, COMSOL.
For the cases on the irregular geometric domain, the mesh should be fine enough to depict the corner accurately. Therefore, both these two stages are time-consuming due to massive iterations \cite{majumdar2001rans,chen2014communication}. Overall, such numerical methods could be applied to temperature field prediction on the irregular geometric domain but with the time-consuming drawback.

Another natural intuition is constructing a surrogate model to learn the mapping from design parameters to temperature fields. Many machine learning (ML) methods have been proven promising for surrogate modeling of PDE-based systems \cite{capuano2019smart,yao2020fea,zhao2021surrogate}. Popularly, ML methods are divided into two classes: data-driven and physics-informed methods. 
Data-driven methods are widely used by training models with labeled data, and many data-driven models are proposed for temperature field prediction. Back in the early 21st century, classic methods, such as Kriging  \cite{hannat2014application}, Radial Basis Function \cite{li2001neuro}, Random Forest \cite{dasari2019random}, were introduced as mainstream surrogate modeling in aircraft design. However, these methods have limitations on solving high-dimensional problems. 
Recently, by taking full advantage of deep learning methods in dealing with high-dimensional problems, many interdisciplinary integration milestone works emerged.
Similarly, deep neural networks (DNNs) have been introduced to serve as regression models by enforcing temperature prediction into an image-to-image regression problem. For instance, the fully connected neural network(FCNN) models \cite{zakeri2019deep}, the fully convolutional neural networks \cite{edalatifar2021using}, the conditional generative adversarial networks \cite{farimani2017deep} and others \cite{chen2021deep}. Especially, Chen et al. \cite{chen2020heat} proposed a new deep learning surrogate-assisted heat source layout optimization method, using the feature pyramid network \cite{lin2017feature} as the surrogate model to evaluate thermal performance accurately under different input layouts. Compared with the FPN model, the U-Net model \cite{ronneberger2015u} also has a fairly good performance. Ma et al. \cite{ma2020combined} used a U-Net model and proposed a combined data-driven and physics-driven method to improve prediction accuracy. Li et al. \cite{li2020fourier} raised a new neural operator by parameterizing the integral kernel directly in Fourier space (FNO), allowing for an expressive and efficient architecture. 
However, since all of the above models predict the entire temperature field, these models are needed to be trained on vast labeled training data. Furthermore, Hei{\ss} et al. \cite{heiss2021neural} and Lye et al. \cite{lye2021multi} proposed a novel neural method for high-dimensional parametric PDEs. They adopt the idea of combining DNNs and Proper Orthogonal Decomposition (POD), which allows an efficient
approximation of the solution. They also present a multi-level scheme to improve the performance of the algorithm and reduce the solution error.

Another branch give birth to physics-informed methods. The perspective is using neural networks as solvers by constraining PDEs into the loss function. When governing PDEs are known, their solutions can be learned in a physics-informed way with little or even no data. A milestone work was developed by Raissi et al. \cite{raissi2019physics}, named physics-informed neural network (PINN), which combined the PDE residual with the loss function of FCNN as a regulative function.        
Furthermore, Gao et al. \cite{gao2021phygeonet} proposed an unsupervised solver based on CNN for solving the PDE on irregular geometric domains. They used a coordinate mapping method to map irregular domains to regular domains and then proposed a physics-constrained CNN learning architecture to learn solutions of parametric PDEs on irregular domains without any labeled data. However, every time the shape changes, the method requires remeshing and retraining to get the physical field.
Besides, motivated by PINN, PDE-based loss function can also be explored for surrogate modeling.
Zhao et al. \cite{zhao2021physics} and Sharma et al. \cite{sharma2018weakly} developed a physics-informed CNN for the thermal surrogate modeling by using the U-Net model, which can learn the mapping from the boundary conditions of heat area to the steady temperature field without label data. The work \cite{zhao2021physics} complied with a method that the heat conduction equation was discretely constructed as a loss function using the finite difference method; the Dirichlet and Neumann boundary conditions were represented as image filling. As a surrogate model, it is challenging to deal with irregular geometric domains because regular geometries with the image-like format are needed.
Whether data-driven or physics-informed surrogate modeling methods are primarily used to deal with rectangular domains, few works have been explored for irregular geometric.

In this paper, for the TFP-IGD problem, motivated by the classic numerical formula, we propose a physics and data co-driven surrogate modeling method. Firstly, to describe irregular geometric regions more efficiently and in a regular image-like format (i.e., rectangular domains with uniform grids).
We parameterize the irregular boundary with a Bezier curve and then use body-fitted coordinate technology to map irregular physical planes to regular computational planes. Secondly, for the sake of overcoming the time-consuming shortcoming of traditional simulation software: we construct the meshing surrogate with nonlinear elliptic equation residuals as the loss function in the mesh generating stage; then, the thermal surrogate is adopted for temperature field prediction in an image format. It is worth noting that, in order to improve the prediction accuracy and reduce the dependence on the dataset, the thermal surrogate adopts a data-driven-based multi-level POD model instead of a graph-to-graph regression model. Finally, the results of the two surrogates are combined to form an irregular temperature field.        

In conclusion, this work develops a novel physics and data co-driven surrogate model, which could reach a fast prediction from geometric shape parameters to temperature field. On the one hand, our method can overcome the large computational burden of traditional numerical methods; on the other hand, and the proposed method can deal with the problem of temperature field prediction in irregular geometric regions that is difficult for CNN to handle. The main contributions could be concluded as follows.
\begin{enumerate}
	\item [(1)] In order to deal with irregular geometric domains, we use the Bezier curve to approximate the irregular area; and apply the body-fitted coordinate technique to complete the mapping from the irregular physical plane to the regular computational plane.
	\item [(2)] We develop a physics-informed network for mesh generation based on nonlinear elliptic equation and finite difference method. The network can be trained by the physics-informed loss with few labeled data and can be applied to construct a structural mesh in various irregular domains. 
	\item [(3)]  A data-driven surrogate model based on the multi-level POD method is proposed to map from geometric parameters to temperature fields. The experiment verifies that the proposed method can accurately predict even with few training samples.
	\item [(4)] We propose a physics and data co-driven surrogate model framework for temperature field prediction on irregular domains. The experiments demonstrate that the temperature field can be accurately and quickly obtained with given geometric parameters. 
\end{enumerate}
The remainder of this paper is structured as follows. Mathematical modeling for irregular temperature field prediction is introduced in Section \ref{sec2}. The physics and data co-driven surrogate model framework is presented in Section \ref{sec3}. Numerical results on heat equations and the comparison of our approach with other methods are presented in Section \ref{sec4}. Finally, Section \ref{sec5} concludes the paper.

\section{Mathematical modeling for temperature field prediction on irregular geometric domain}\label{sec2}

\begin{figure*}[htb]
	\centering
	{\includegraphics[scale=0.7]{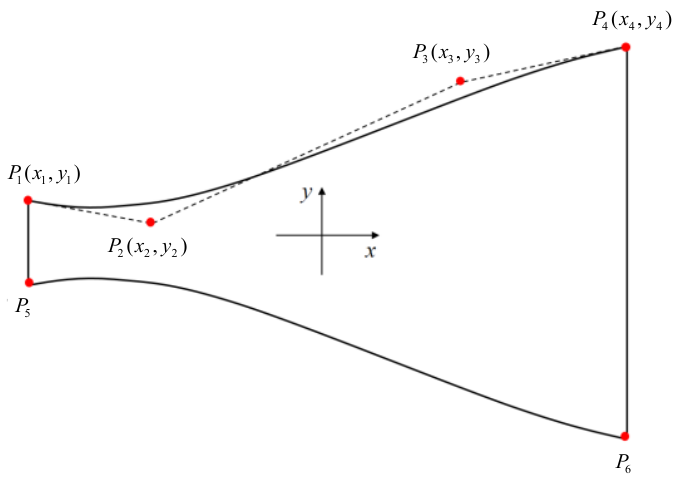}}
	\caption{The illustration of the irregular geometric domain.}\label{igd}
\end{figure*}

In the aircraft thermal structure design, computing the temperature field with a large variety of geometric shapes is required. Each shape modification based on the surrogate modeling means a new input and corresponding prediction followed. 
In this section, we first define a temperature field prediction problem on an irregular geometric domain.
In a two-dimensional plane, the steady-state temperature field without internal heat source can be computed by solving the Laplace equation,
\begin{equation}\label{3}
\begin{aligned}
& \frac{\partial^{2} T}{\partial x^{2}}+\frac{\partial^{2} T}{\partial y^{2}}=0, \\
& \text{Boundary}: \quad T=T_{b}. \\
\end{aligned}
\end{equation}
The Dirichlet boundary condition is given for the convenience of explaining the proposed approach, and $T_{b}$ represents temperature values on different curves. 
Figure \ref{igd} shows one irregular quadrangle, where geometry shape of the irregular domain is controlled by curves $P_1P_4$ and $P_5P_6$. 
Due to the wide variation range of curve $P_1P_4$ and $P_5P_6$, it is challenging to predict the temperature field under different irregular geometric domains accurately. Intuitively, the domain could be more complex.

Motivated by curve fitting, an irregular geometric boundary can be represented by geometric parameters, which can simplify the representation of the boundary curve. In this paper, the Bezier curve is chosen as the boundary curve of irregular geometric domains, which is extensively utilized in the geometry design of aircraft and automobile fields. Once the control vertex $P_i(x_i,y_i) \left( i=0,1,\cdots,n \right)$ is fixed, the parametric equation of the n-order Bezier curve segment can be expressed as follows:
\begin{equation}\label{1}
B(t)=\sum_{i=0}^{n} B_{n, i}(t) P_{i}(x_i, y_i),
\end{equation}
where $B_{n, i}(t)$ is the Bernstein base function, has the following form:
\begin{equation}\label{2}
B_{n, i}=C_{n}^{i}t^{i}(1-t)^{n-i}.
\end{equation}

Specifically, Figure \ref{igd} also shows a schematic depiction of the irregular geometric domain fitted by a cubic Bezier curve with four control vertices. In this case, we make an assumption that this domain is symmetric as the x-axis.  
We set $P_1,P_2,P_3,P_4$ are the four control vertices of the cubic Bezier curve $P_1P_4$, which is similar for curve $P_5P_6$.

For solving such a problem, the coordinates of these control vertices are considered as input parameters. Consequently, our goal is translated into constructing an end-to-end surrogate model, which can efficiently learn the mapping from the given geometric parameters to the temperature field.

\section{A physics and data co-driven surrogate model for TFP-IGD}\label{sec3}
\begin{figure*}[htb]
	\centering
	{\includegraphics[scale=0.7]{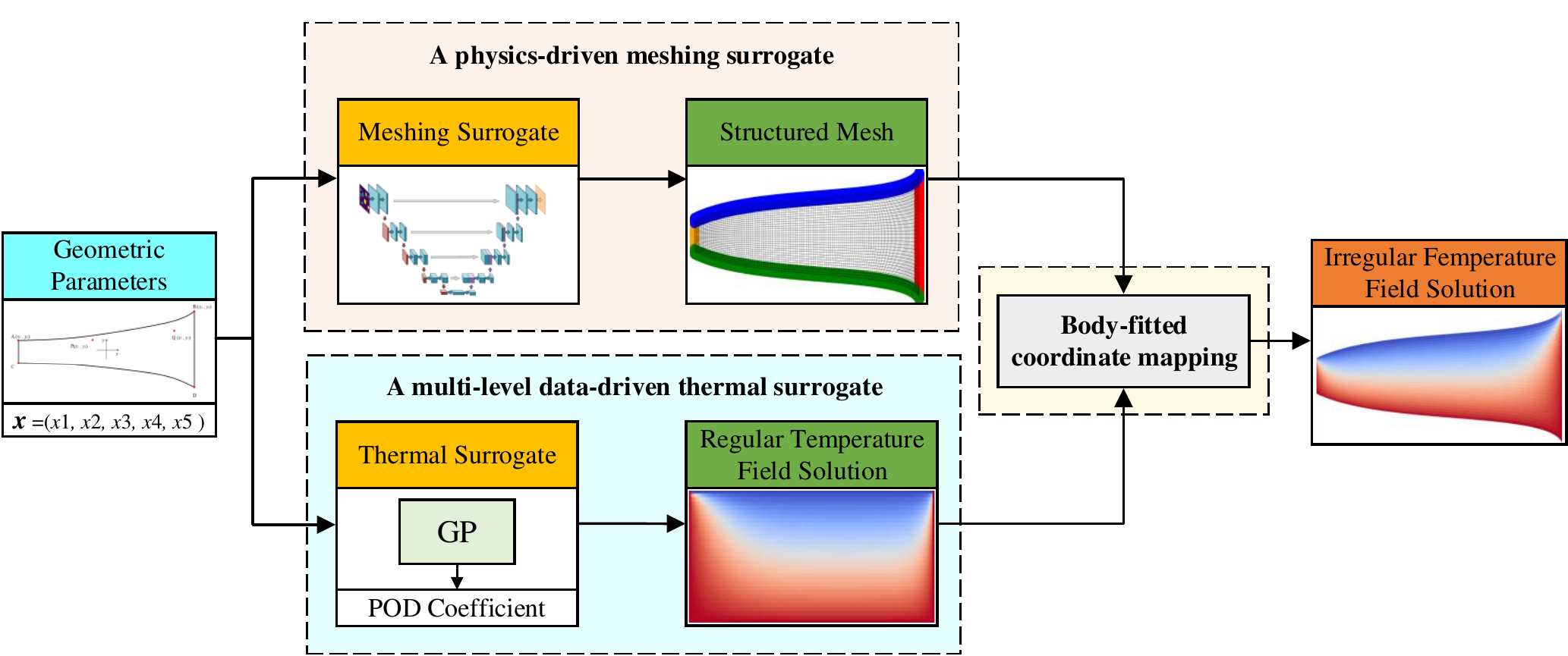}}
	\caption{The framework of the physics and data co-driven surrogate.}\label{framework}
\end{figure*}

In this section, the physics and data co-driven surrogate modeling framework is presented for TFP-IGD. Motivated by the classic numerical analysis formula, the framework has been translated into a two-stage model: the meshing surrogate for structural mesh generating and the thermal surrogate for predicting the regular temperature field mapping from the irregular domain. The meshing surrogate is implemented by constructing a PDE-based loss function. In contrast to the end-to-end approach, by adopting the proper orthogonal decomposition (POD) technology, the thermal surrogate is used to predict the POD coefficients of fixed temperature field basis rather than the complete temperature field. Figure \ref{framework} illustrates the proposed physics and data co-driven surrogate model framework.
As shown in Figure \ref{framework}, with each afferent irregular geometric boundary, we first input the boundary into the meshing surrogate for meshing and obtain the coordinates of all mesh points. Then parameterize the boundary and put it into the thermal surrogate to obtain the mapped regular temperature field. Finally, the irregular temperature field is obtained by combining the grid position information in the meshing surrogate and the scalar temperature field information in the thermal surrogate.

It is organized as follows. In Section \ref{sec31}, the body-fitted coordinate mapping is introduced. Section \ref{sec32} presents a physics-driven surrogate model for the meshing surrogate, and Section \ref{sec33} describes a multi-level data-driven method for the thermal surrogate.

\subsection{Body-fitted coordinate mapping}\label{sec31}
\begin{figure*}[htb]
	\centering
	{\includegraphics[scale=0.8]{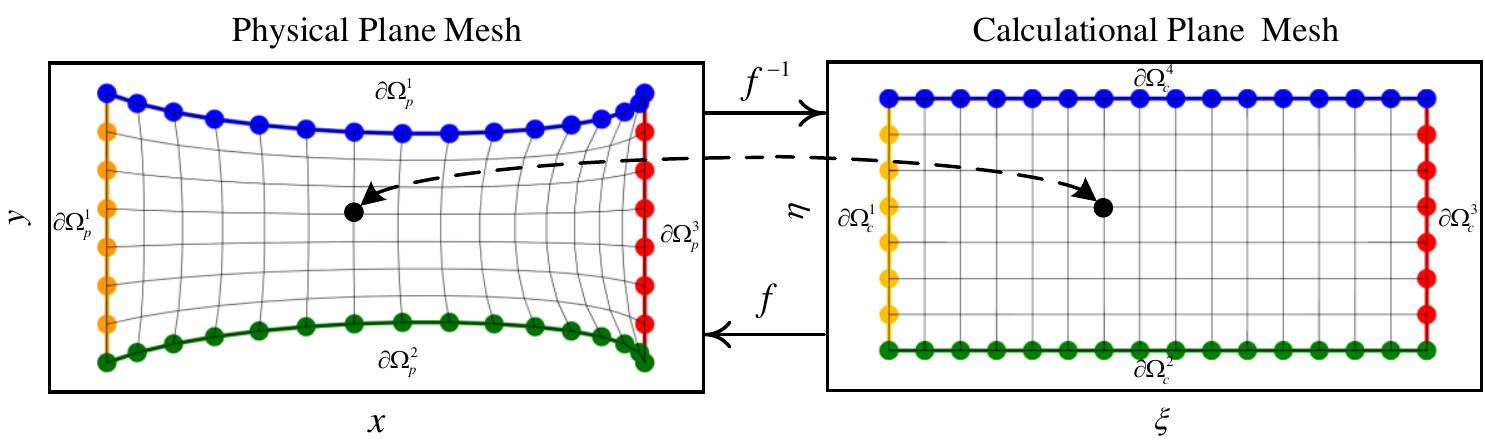}}
	\caption{A schematic diagram of coordinate mapping between irregular physical and computational plane.}\label{map}
\end{figure*}

As shown in Figure \ref{map}, irregular grid points generally appear on the physical plane, making it difficult to predict the temperature field with the image format. In order to overcome this tricky issue, body-fitted coordinate systems \cite{thompson1982boundary} were proposed. The key idea of the body-fitted coordinate systems is utilizing coordinate transformation techniques to map solution fields from the irregular physical plane to the rectangular calculation plane. (as shown in Figure \ref{map}).

The forward/inverse mapping between coordinates of the irregular physical plane ($\Omega_{p}$) and regular computational plane ($\Omega_{c}$) can be defined as,
\begin{equation}
\mathbf{x}=\mathbf{f}(\bm{\xi}), \quad \bm{\xi}=\mathbf{f}^{-1}(\mathbf{x}),
\end{equation}
where $\mathbf{f}: \Omega_{c} \mapsto \Omega_{p}$ denotes the forward map, and $\mathbf{f}^{-1}: \Omega_{p} \mapsto \Omega_{c}$ denotes the inverse map; $\mathbf{x} \doteq [x, y] \in \Omega_{p}$ and $\bm{\xi} \doteq[\xi, \eta] \in \Omega_{c}$ represent mesh coordinates of the physical plane and computational plane, respectively. 
In our work, elliptic coordinate mapping is introduced to complete the coordinate transformation in Figure \ref{map}. 
The generation of body-fitted coordinate can be treated as solving the differential equation with the Dirichlet boundary condition. Motivated by the work of \cite{thompson1974automatic}, elliptic equations with Dirichlet boundary condition can be applied to approximate the mapping. For example, the mapping can be obtained by solving a diffusion equation as follows:
\begin{equation}\label{7}
\begin{aligned}
\nabla^{2} \bm{\xi}(\mathbf{x})=\xi_{xx}+\xi_{yy}=0, \\
\nabla^{2} \bm{\eta}(\mathbf{x})=\eta_{xx}+\eta_{yy}=0,
\end{aligned}
\end{equation}
with the boundary conditions:
\begin{equation}
\begin{aligned}
\bm{\xi}(\mathbf{x})=\bm{\xi}_{b}, \quad \text{for}~~\forall~\mathbf{x} \in \partial \Omega_{p}^{i}, \quad i=1, \cdots, 4.
\end{aligned}
\end{equation}
The coordinates of the physical plane and computational plane are denoted by $\mathbf{x} \doteq[x, y]$ and $\bm{\xi} \doteq[\xi, \eta]$. However, considering that the area on the physical plane is invariably irregular, it is challenging to solve (\ref{7}). By interchanging the independent and dependent variables in (\ref{7}), the following diffusion equations corresponding to the calculation plane in terms of physical coordinate $x,y$ is derived,
\begin{equation}\label{9}
\begin{aligned}
&\alpha \frac{\partial^{2} x}{\partial \xi^{2}}-2 \beta \frac{\partial^{2} x}{\partial \xi \partial \eta}+\gamma \frac{\partial^{2} x}{\partial \eta^{2}}=0, \\
&\alpha \frac{\partial^{2} y}{\partial \xi^{2}}-2 \beta \frac{\partial^{2} y}{\partial \xi \partial \eta}+\gamma \frac{\partial^{2} y}{\partial \eta^{2}}=0,
\end{aligned}
\end{equation} 
with the boundary conditions:
\begin{equation}
\begin{aligned}\label{bc}
\mathbf{x}(\boldsymbol{\xi})=\mathbf{x}_{b}, \quad  &\text{for}~~\forall~\bm{\xi} \in \partial \Omega_{c}^{i}, \quad i=1, \cdots, 4, 
\end{aligned}
\end{equation}
and $\alpha,\beta,\gamma$ are given by
\begin{equation}
\begin{aligned}\label{11}
&\alpha=\left(\frac{\partial x}{\partial \eta}\right)^{2}+\left(\frac{\partial y}{\partial \eta}\right)^{2}, \\
&\gamma=\left(\frac{\partial x}{\partial \xi}\right)^{2}+\left(\frac{\partial y}{\partial \xi}\right)^{2}, \\
&\beta=\frac{\partial x}{\partial \xi} \frac{\partial x}{\partial \eta}+\frac{\partial y}{\partial \xi} \frac{\partial y}{\partial \eta}.
\end{aligned}
\end{equation}
By solving (\ref{9}) with boundary condition defined in (\ref{bc}) numerically (i.e., iterative method), the discrete values of the forward map $f$ are obtained.

\subsection{A physics-driven meshing surrogate}\label{sec32}
It is very time-consuming to use numerical methods to complete the structural meshing for involving multiple iterative calculations.
Therefore, a physics-driven approach for the meshing surrogate is introduced in this part. We first introduce the finite difference discretization of the governing equations, then we use it to build a physics-informed loss for the meshing surrogate.
\subsubsection{The finite difference discretization of the governing equations}\label{sec321}
In order to transform the differential equation residuals into loss function in the deep learning model, we employ the finite difference method to obtain the discrete values of (\ref{9}). 
\begin{figure*}[htb]
	\centering
	{\includegraphics[scale=0.5]{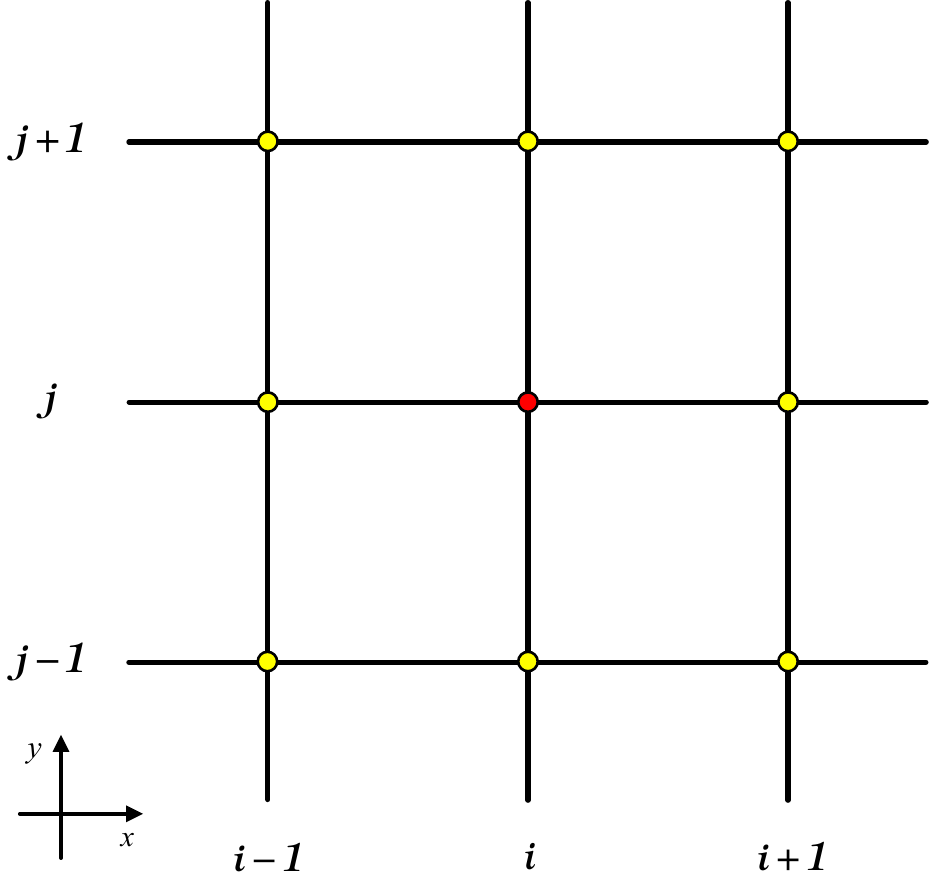}}
	\caption{Illustration of mesh point in the regular domain.}\label{cdm}
\end{figure*}
More specifically, the discretization of this nonlinear system of equations is accomplished by using the central difference method. As shown in Figure \ref{cdm}, taking the $x$ direction as an example, the first-order derivative of the coordinate system is discrete as follows,
\begin{equation}
\begin{aligned}
&\frac{\partial x}{\partial \xi}\approx\frac{x_{i+1, j}-x_{i-1, j}}{2} ,\\
&\frac{\partial x}{\partial \eta}\approx\frac{x_{i, j+1}-x_{i, j-1}}{2} .
\end{aligned}
\end{equation}
Then, the second-order derivative can be approximated by
\begin{equation}\label{12}
\begin{aligned}
&\frac{\partial^{2} x}{\partial \xi^{2}}\approx x_{i+1, j}+x_{i-1, j}-2 x_{i, j}, \\
&\frac{\partial^{2} x}{\partial \eta^{2}}\approx x_{i, j+1}+x_{i, j-1}-2 x_{i, j}, \\
&\frac{\partial^{2} x}{\partial \xi \partial \eta}\approx\frac{x_{i+1, j+1}+x_{i-1, j-1}-x_{i+1, j-1}-x_{i-1, j+1}}{4}.
\end{aligned}
\end{equation}
Similarly, second-order derivative in the $y$ direction is available as
\begin{equation}\label{13}
\begin{aligned}
&\frac{\partial^{2} y}{\partial \xi^{2}}\approx y_{i+1, j}+y_{i-1, j}-2 y_{i, j}, \\
&\frac{\partial^{2} y}{\partial \eta^{2}}\approx y_{i, j+1}+y_{i, j-1}-2 y_{i, j}, \\
&\frac{\partial^{2} y}{\partial \xi \partial \eta}\approx\frac{y_{i+1, j+1}+y_{i-1, j-1}-y_{i+1, j-1}-y_{i-1, j+1}}{4}.
\end{aligned}
\end{equation}
In addition, $\alpha,\beta$ and $\gamma$ can also be gained, 
\begin{equation}
\begin{aligned}\label{14}
&\alpha=\left(\frac{\partial x}{\partial \eta}\right)^{2}+\left(\frac{\partial y}{\partial \eta}\right)^{2}\approx\left(\frac{x_{i, j+1}-x_{i, j-1}}{2}\right)^{2}+\left(\frac{y_{i, j+1}-y_{i, j-1}}{2}\right)^{2}, \\
&\beta=\frac{\partial x}{\partial \xi} \frac{\partial x}{\partial \eta}+\frac{\partial y}{\partial \xi} \frac{\partial y}{\partial \eta}\approx\left(\frac{x_{i+1, j}-x_{i-1, j}}{2}\right)\left(\frac{x_{i, j+1}-x_{i, j-1}}{2}\right)+\left(\frac{y_{i+1, j}-y_{i-1, j}}{2}\right)\left(\frac{y_{i, j+1}-y_{i, j-1}}{2}\right), \\
&\gamma=\left(\frac{\partial x}{\partial \xi}\right)^{2}+\left(\frac{\partial y}{\partial \xi}\right)^{2}\approx\left(\frac{x_{i+1, j}-x_{i-1, j}}{2}\right)^{2}+\left(\frac{y_{i+1, j}-y_{i-1, j}}{2}\right)^{2}.
\end{aligned}
\end{equation}
By combining (\ref{12}), (\ref{13}), and (\ref{14}), the discretization of (\ref{9}) is completed. Further, (\ref{9}) can be integrated into the loss function of the deep learning model to form a physics-driven framework in Section \ref{sec322}.
\subsubsection{Physics-driven surrogate model for fast meshing}\label{sec322}
\begin{figure*}[htb]
	\centering
	{\includegraphics[scale=0.7]{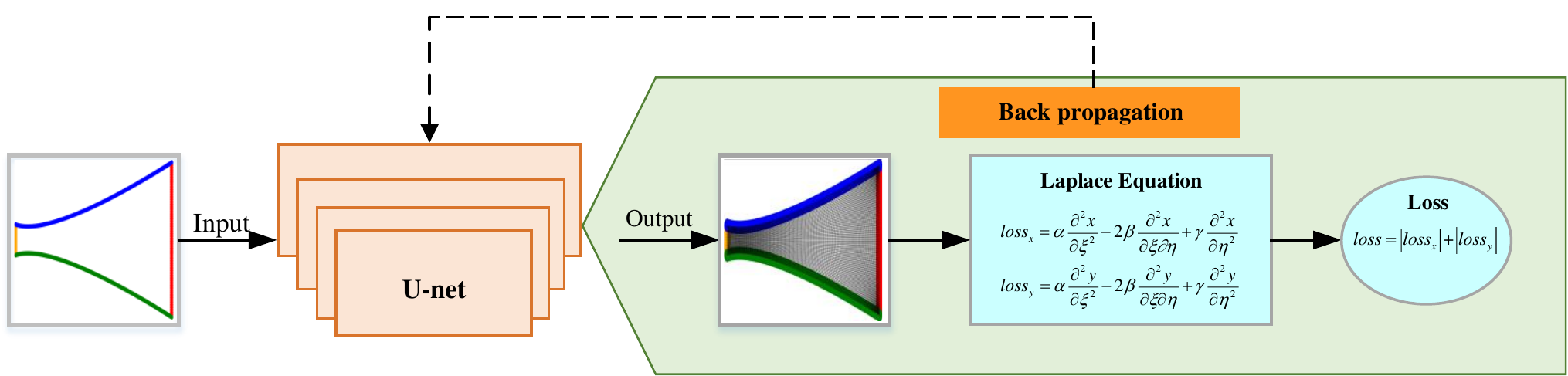}}
	\caption{The illustration of fast meshing method based on physics-driven method. }\label{frameunet}
\end{figure*}

Based on (\ref{12}), (\ref{13}), and (\ref{14}), the structured mesh can be generated through iterative methods. However, the disadvantage is that generating structured mesh would be time-consuming, especially when the number of grid points is burgeoning.
To improve the efficiency of mesh generation, a novel physics-constrained CNN learning architecture has been proposed without any labeled data. The loss function of this framework is set as
\begin{equation}
\begin{aligned}\label{16}
&\operatorname{loss}_{x}=\alpha \frac{\partial^{2} x}{\partial \xi^{2}}-2 \beta \frac{\partial^{2} x}{\partial \xi \partial \eta}+\gamma \frac{\partial^{2} x}{\partial \eta^{2}}, \\
&\operatorname{loss}_{y}=\alpha \frac{\partial^{2} y}{\partial \xi^{2}}-2 \beta \frac{\partial^{2} y}{\partial \xi \partial \eta}+\gamma \frac{\partial^{2} y}{\partial \eta^{2}}, \\
&\operatorname{loss}=\left| \operatorname{loss}_{x} \right|+\left| \operatorname{loss}_{y} \right|.
\end{aligned}
\end{equation}
Combining formula (\ref{12}),(\ref{13}), and (\ref{14}), we can get the discrete values of formula (\ref{16}). In this way, we can get the loss function of the network. Then we can train the CNN based upon backpropagation. Figure \ref{frameunet} states the framework of physics-constrained CNN learning.
We encode the coordinates of grid points $(x, y)$ on irregular domains into a two-channel image format. The input of the network is a two-channel image containing only boundary values, where the rest of the image is filled with value $0$. The U-net model acts as a mesh generator, which generates coordinates corresponding to grid points on different irregular geometric domains. Similarly, the output of the network is also a two-channel image, which represents coordinates of boundary and internal grid points $(x, y)$ respectively.

\subsection{A multi-level data-driven thermal surrogate}\label{sec33}

Using numerical methods to solve the temperature field contains multiple iteration calculations, making the solution time-consuming. Therefore, a multi-level data-driven approach for temperature field prediction is introduced in this part. In order to facilitate the subsequent optimization process, we use the geometric parameters as the model input and adopt the multi-level POD model with the POD coefficients as output. Then the temperature field can be obtained. It is organized as follows. We first introduce two key techniques for the multi-level data-driven model: The numerical method under the multi-level framework and the POD technique; then, we describe the multi-level data-driven model for the regular temperature field after mapping.
\subsubsection{Numerical method under the multi-level framework}\label{sec331}

\begin{figure*}[htb]
	\centering
	{\includegraphics[scale=0.75]{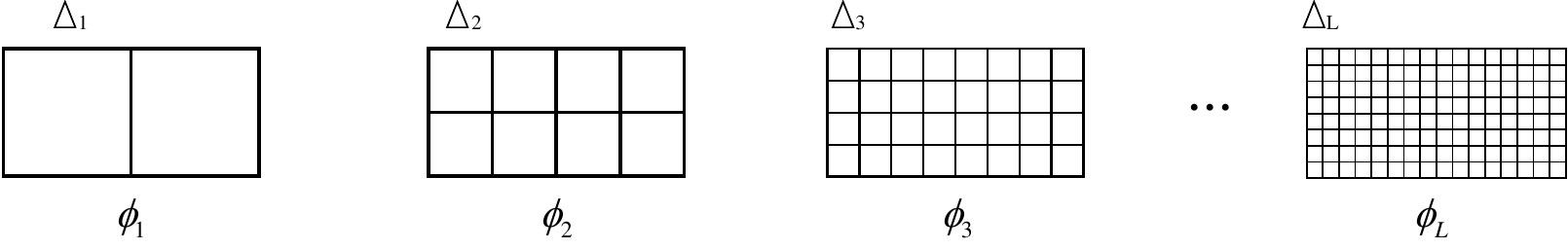}}
	\caption{A schematic for a sequence of nested grids in multi-stage finite difference method.}\label{mesh}
\end{figure*}

Motivated by the multi-grid technique, the numerical method under the multi-level framework is proposed \cite{heiss2021neural}. For the sake of making the prediction more precise, integrating into the idea of multi-grid, the complicated prediction process has been divided into numerous simpler sub-processes. We define $\phi_{1} \subset \cdots \subset \phi_{L} \subset M(D)$ as a nested grid space obtained by successive refinement on the basis of coarse grid over the spatial domain $D$, as shown in Figure \ref{mesh}, with $v_{i}, i=1,\cdots,l$ being obtained by the numerical solution, such as FDM, FEM, and FVM.
The $v_L$ is the finest solution. 

\begin{figure*}[htb]
	\centering
	\subfigure [The subtraction operation of getting the residuals]
	{\includegraphics[scale=1]{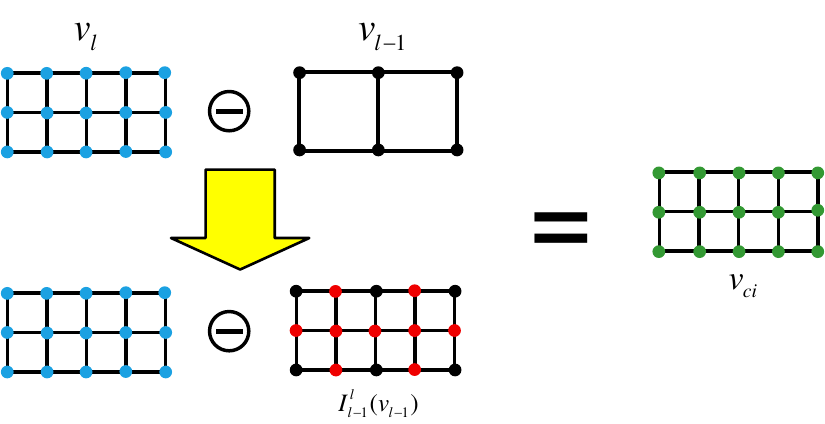}\label{sub}}
	\subfigure [The addition operation between residuals]
	{\includegraphics[scale=0.75]{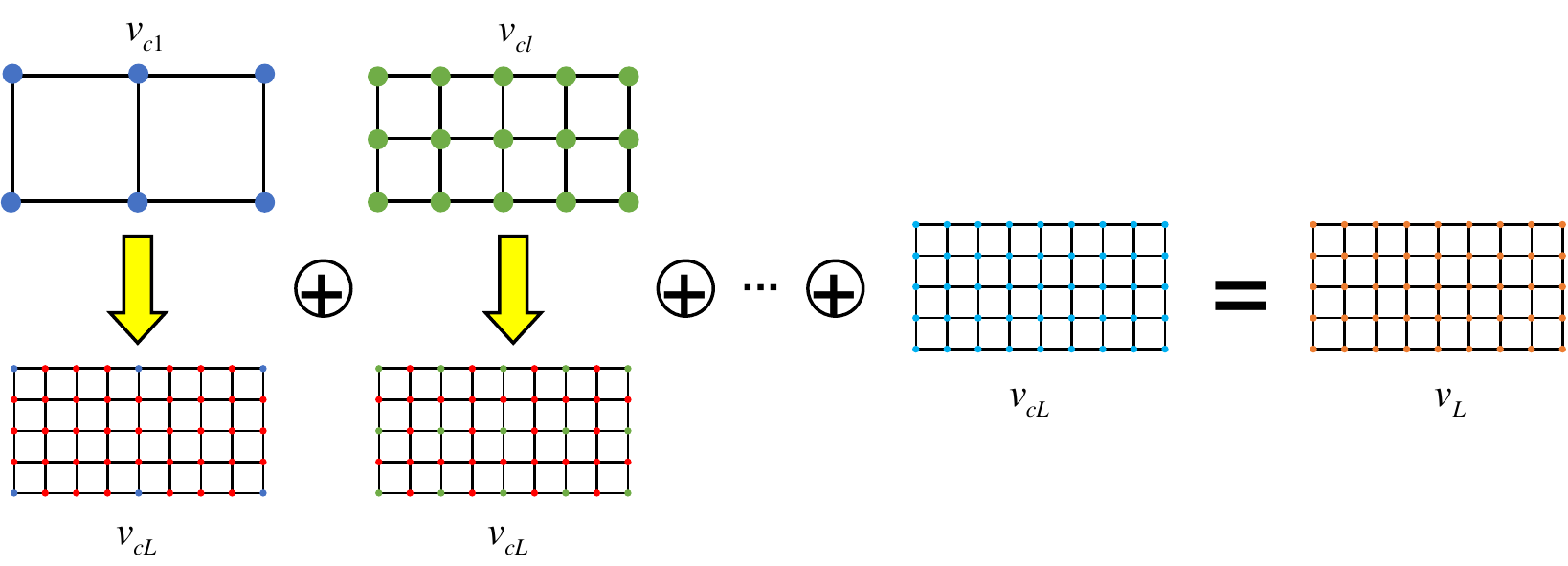}\label{add}}
	\caption{The schematic diagram of two operations in the multi-stage finite difference method. The dots represent the temperature values on the grid points, the red represents the results obtained by interpolation, and the other colors represent the previously calculated temperature values of the grid points with different precisions.}\label{mfdm}
\end{figure*}

Thus, we can bring the modified form of the solution $v_L$ on the most refined mesh $\phi_{L}$ as
\begin{equation}\label{18}
v_{L}^{'}=v_{1}+\sum_{l=2}^{L} (v_{l}-v_{l-1})=v_{1}+\sum_{l=2}^{L} v_{c l}=\sum_{l=1}^{L} \widetilde{v_{l}}.
\end{equation}
As can be seen from (\ref{18}), the computation of the approximate solution for the finest mesh consists of two operations: the first is a subtraction operation to get the residuals as in Figure \ref{sub}, and the second is an addition operation between residuals like in Figure \ref{add}. For the subtraction operation, we define $I_{l-1}^{l}$ as the transfer operator from sparse mesh ($\phi_{l-1}$) to the precise mesh ($\phi_{l}$), then the residual can be expressed as
\begin{equation}
v_{c l}=v_{l}-v_{l-1}=v_{l}-I_{l-1}^{l}\left(v_{l-1}\right) \quad l=2, \cdots, L.
\end{equation}
In fact, $I_{l-1}^{l}$ can be realized in many ways, such as interpolation. Equally, the addition operation can be approached similarly. 
In this way, we transform the finest mesh solution $v_{L}^{'}$ into a combination of sub-process solutions by using the multi-stage finite difference method.

\subsubsection{POD technology}\label{sec332}
Then, the POD technology will be adopted to reduce the order of each sub-process solution. From Figure \ref{mesh}, we can apparently see that dimension $\phi_{l}$ grows exponentially with the level $l$. Aiming to solve such high-dimensional problem, the POD technology is adopted to generate a basis space $V_{l}=\operatorname{span}\left\{v_{l}^{1}, \ldots, v_{l}^{\operatorname{dim} V_{l}}\right\}$ to represent approximations of $\tilde{v_{l}}$ 
in (\ref{18}) in an efficient way. The final solution $v_{L}$ consists of many subprocesses $\tilde{v_{l}}$.
\begin{equation}
\tilde{v_{l}} = \sum_{i=1}^{\operatorname{dim} V_{l}} c_{l}^{i} v_{l}^{i},
\end{equation}
where $c_{l}^{i}$ are the POD coefficients. 

In this way, the solution $\tilde{v_{l}}$ could be decomposed as a linear combination of $c_{l}^{i}$ and $v_{l}^{i} $. Then given the POD basis $V_{l}$, only POD coefficients $c_{l}^{i}$ need to be predicted. Due to the mapping between PDE parameters and POD coefficients, the model training process is simplified and time-saving.

\subsubsection{ A multi-level data-driven method for the thermal surrogate}\label{sec333}

\begin{figure*}[htb]
	\centering
	{\includegraphics[scale=0.7]{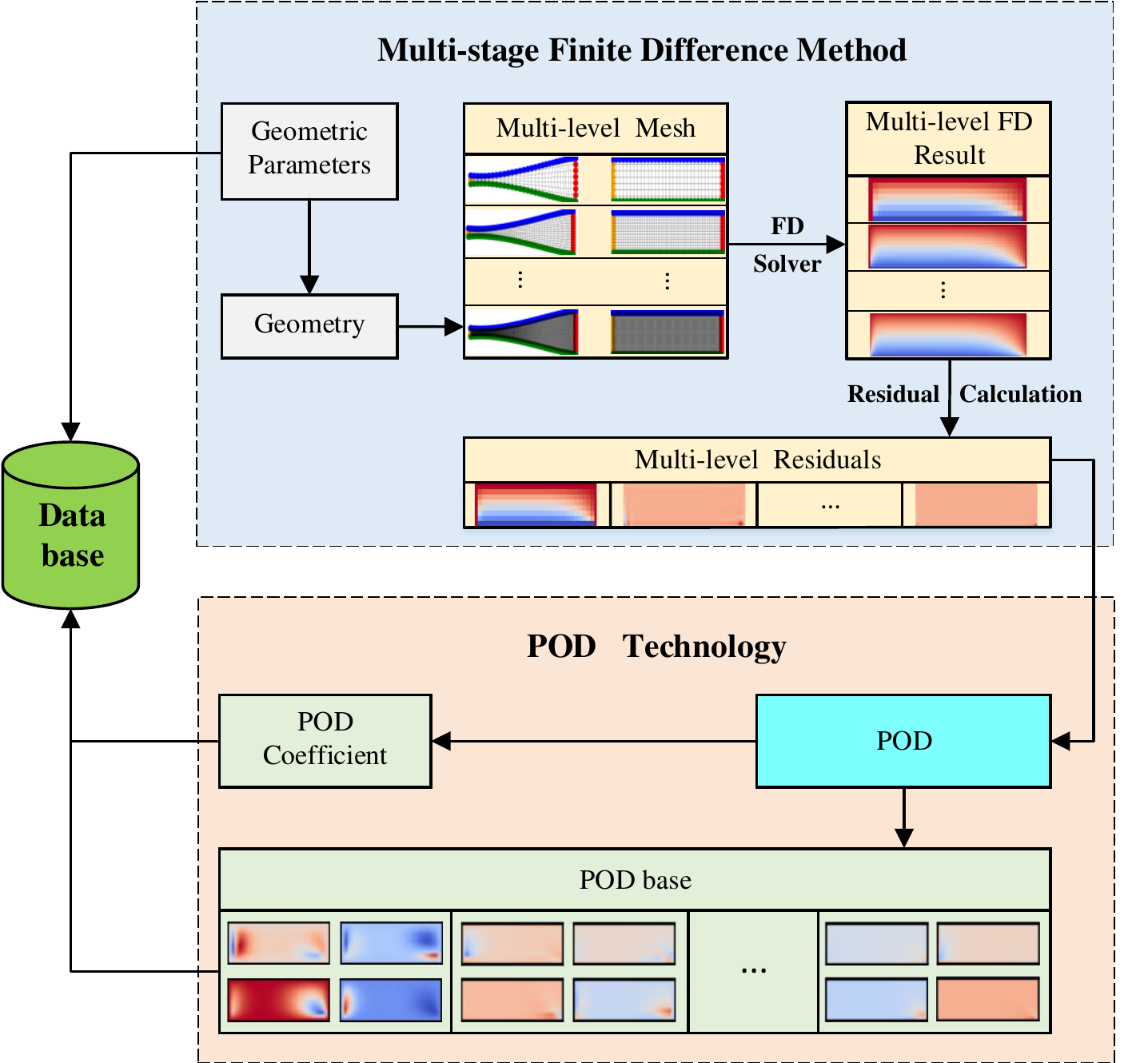}}
	\caption{The procedure of data generation.}\label{data}
\end{figure*}

\paragraph{\textbf{Data preparation}}
Different from other end-to-end data-driven methods, our method constructs a mapping between geometric parameters (described in Section \ref{sec2}) and POD coefficients (described in Section \ref{sec332}). 
To get the multi-level POD coefficients, two essential procedures need to be completed during the data preparation. Figure \ref{data} illustrates the whole process.

First, we use the multi-stage finite difference method to generate finite-difference decomposition of different mesh precision and compute the residuals between them as in Section \ref{sec331}. Then, the POD technology has been used to reduce the order of residuals at different grid precision in Section \ref{sec332}. In this way, we can get the multi-level POD coefficients and complete the data preparation.

\paragraph{\textbf{Model training and prediction}} 
As mentioned earlier, the basis and coefficients are obtained by the POD technique. In this way, a data-driven surrogate model can be constructed to map from geometric parameters to the POD coefficients. Then the temperature field can be obtained by the inner product between the POD coefficients and the POD basis. Since both geometric parameters and coefficients are low dimensions, many machine learning models could be capable, such as DNNs, Random forest regression, Gaussian process regression, etc.

In our multi-level data-driven surrogate model, we use the Gaussian process (GP) regression model: $\mathbb{R}^{p} \rightarrow \mathbb{R}^{\operatorname{dim} V_{l}}$ to learn the basis coefficients $c_{l}^{i}$ ($p$ is the dimension of the input geometry parameters), 
\begin{equation}
v_{L}^{'}=\sum_{l=1}^{L} \tilde{v_{l}} \approx \sum_{l=1}^{L} \sum_{i=1}^{\operatorname{dim} V_{l}} c_{l}^{i} v_{l}^{i} \approx \sum_{l=1}^{L} \sum_{i=1}^{\operatorname{dim} V_{l}}[G P]_{l} v_{l}^{i}.
\end{equation}
In addition, since it is a coefficient-to-coefficient mapping rather than predicting the entire temperature field, the whole training process will be easily and fast predictable. 

In general, this is a quite concise data-driven framework. As long as the geometric parameters representing the irregular geometric domain are input, the corresponding temperature field under given boundary conditions can be obtained.

\section{Numerical results}\label{sec4}
In this section, Firstly, we evaluate the performance of the meshing surrogate on the prediction accuracy. And to prove the validity of the meshing surrogate, we compare the mesh quality of the physics-driven and data-driven method on U-net with the same structure. Secondly, the experiments for TFP-IGD based on the thermal surrogate are explored. We evaluate the irregular temperature field results, which are obtained by the combination of the meshing surrogate and thermal surrogate. In order to validate the efficacy of the thermal surrogate, we compare the result with some new combine models with the thermal surrogate that have been changed to other models, such as the single-level model, U-net, and FNO model (see Section \ref{sec422}).

In the experiments, mean absolute error (MAE) and mean relative error (MRE) are adopted as the metrics to evaluate the performance of the proposed combined model, where the results solved by finite difference method (FD) are set as the benchmark. In terms of dataset generation (describe in Section \ref{sec333}), 2000 samples of different irregular geometry are generated by the Latin Hypercube sampling method. Among them, $70\%$ of the samples are used for training, and the remaining $30\%$ for testing. We set the mesh series as $L=4$, and the finest mesh size is $64 \times 256$ while the coarsest mesh size is $8 \times 32$.

\subsection{Performance of mesh generation by the meshing surrogate}\label{sec41}
In our study, the U-net model has been adopted in the meshing surrogate model, and Adam is chosen as the optimizer. For the training process, the training epoch has been set to 1000 with batch size 16, and the adaptive learning rate strategy has been applied. A data-driven model will be used for comparison, and the result shows that the meshing surrogate can generate high-quality meshes like numerical methods.

\subsubsection{Prediction performance}
\begin{figure}
	\centering
	{\includegraphics[scale=0.7]{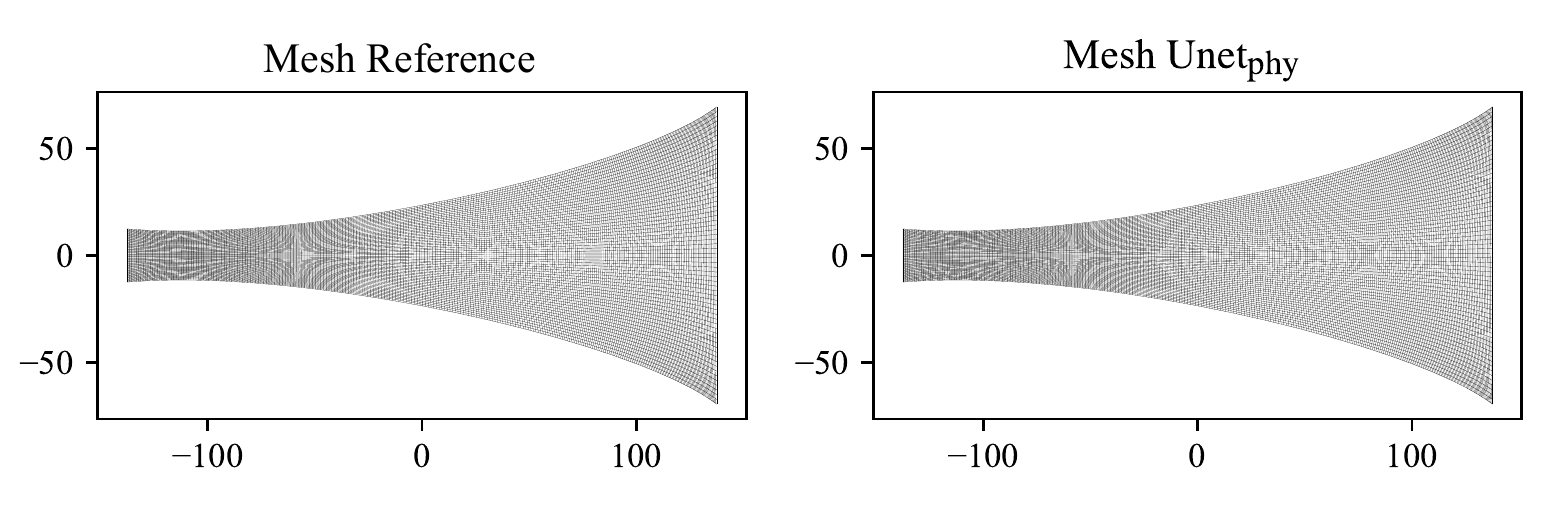}}
	\caption{An illustration of the overall mesh on two test samples."Mesh Reference" represents the mesh calculated by the numerical method, and "Mesh $Unet_{phy}$" represents the mesh obtained by the meshing surrogate.}\label{meshtotal}
\end{figure}

We first train a U-Net model on the training set, denoted by $Unet_{phy}$. Then, we evaluate $Unet_{phy}$ on the test set. Taking the $x$ and $y$ coordinates of the curve boundary as input and through $Unet_{phy}$, we get the $x$ and $y$ coordinates of all grid points as the output of $Unet_{phy}$. After getting the prediction results, the overall grid is obtained by combining the $x$ and $y$ coordinates of each grid point as in Figure \ref{meshtotal}. 
\begin{figure*}[htb]
	\centering
	{\includegraphics[scale=0.7]{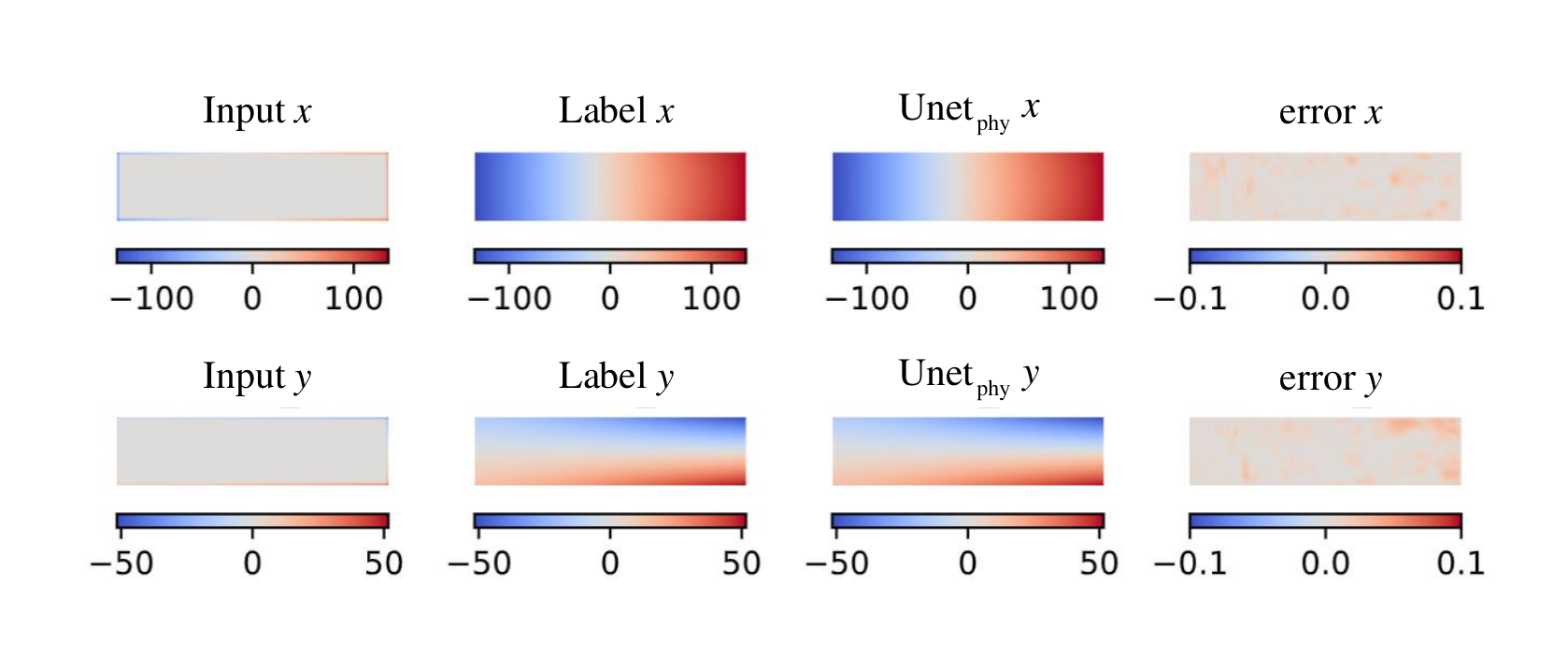}}
	\caption{An illustration of the prediction performance of $Unet_{phy}$ on one test sample.}\label{meshnet}
\end{figure*} 

To better illustrate the prediction accuracy of the meshing surrogate, the MRE in $x$ and $y$ directions are also used as the evaluation metric apart from the MAE, which is called $MRE_x$ and $MRE_y$ separately. $Unet_{phy}$ maintains the MAE of 0.023$K$, while $MRE_x$ and $MRE_y$ are 0.11$\%$ and 0.25$\%$ respectively. As shown in Figure \ref{meshnet}, we take one sample in the test set to deliver the performance. We display the input $x$ and $y$ coordinates, the predicted $x$ and $y$ coordinates, the ground-truth $x$ and $y$ coordinates, and prediction error. It can be seen that $Unet_{phy}$ has a very high prediction accuracy on both $x$ and $y$ coordinates. Based on the three evaluation metrics and Figure \ref{meshnet}, the physics-driven model performs well. However, it is worth noting that the overall mesh result depends not only on the error of the two coordinate axes alone but also depends on the overall mesh quality because it is entirely possible to generate a chaotic mesh. Therefore, we also analyzed the quality of grid generation from the perspective of a structured grid.
Figure \ref{meshtotal} shows the general mesh predicted by the above sample on the test dataset and the ground-truth mesh.
%

\begin{figure*}[htbp]
	\centering
	\vspace{-0.4cm}
	{\includegraphics[scale=0.7]{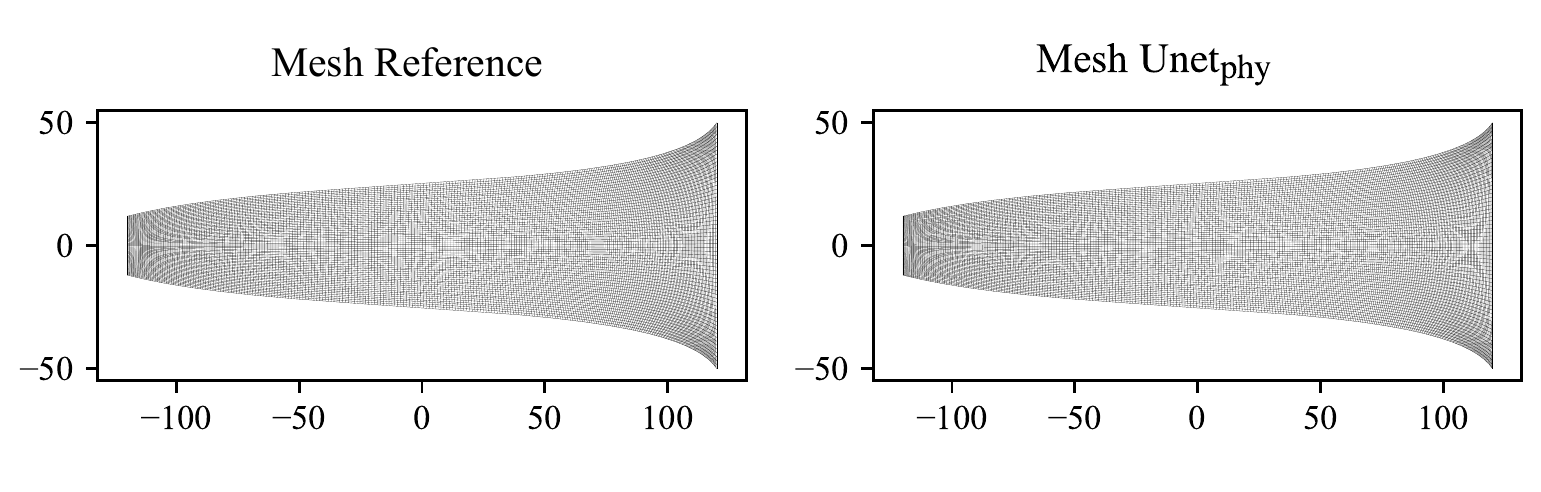}}
	\caption{An illustration of the overall mesh on unseen data that geometric parameters outside the test set range. "Mesh Reference" represents the mesh calculated by the numerical method, and "Mesh $Unet_{phy}$" represents the mesh obtained by the meshing surrogate.}\label{meshtotalunseen}
\end{figure*}

As previously mentioned, the physics-driven model has promising results on the test dataset. Besides, we also found that our method can also obtain high-quality meshes when selecting geometric parameters outside the test set range for mesh generation. The model performs on geometric parameters outside the test set are shown in Figure \ref{meshtotalunseen}.
Since the physics-driven model involves the discretized mesh control equation into the loss function and can learn the relevant information of the mesh control equation, which has strong generalization performance and will have a good performance on geometric parameters outside the test set range either. Whether on the test set or outside the test set, the $Unet_{phy}$ performs admirably.

\subsubsection{Comparison with data-driven method}
For structural mesh generating, a data-driven approach is used in this experiment as a comparison denoted by $Unet_{data}$, which takes boundary points as input and mesh points as output the same as the physics-driven method. Using the U-net model equally while the loss function turns into L2 loss, the label data of actual mesh points are obtained by the iterative approach.

\begin{figure*}[ht]
	\centering
	\vspace{-0.2cm}
	{\includegraphics[scale=0.75]{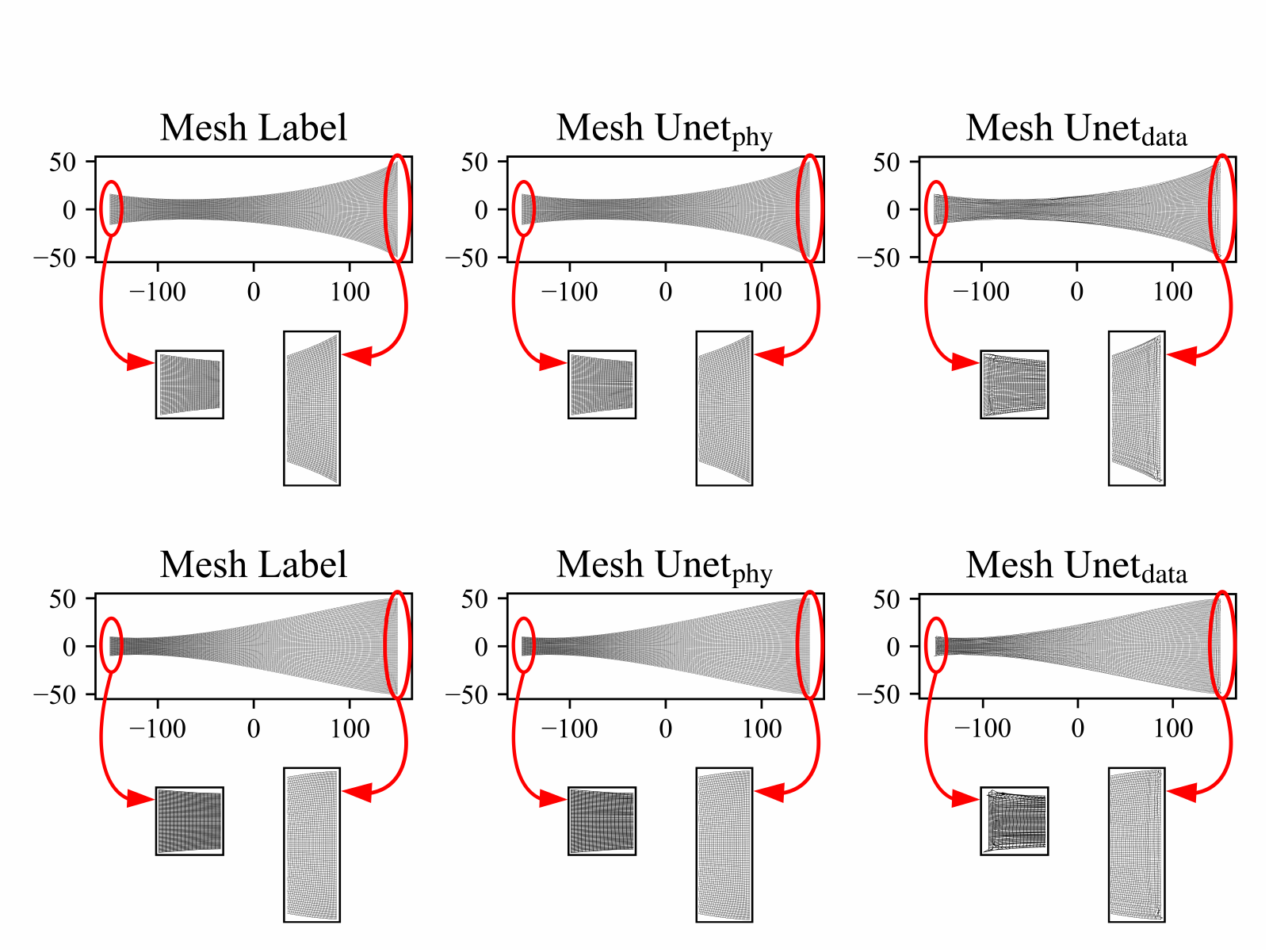}}
	\caption{The comparison between the data-driven and physics-driven method.}\label{meshcompare}
\end{figure*}

The MAE of $Unet_{data}$ on the test set after training is 0.029$K$, which has little difference compared to $Unet_{phy}$, indicating that the data-driven method can also learn the actual mesh points. However, the data-driven method is more like a black-box model, which can only predict the coordinates of each grid point independently. In other words, it is unable to add the connections between grid points into the model, which will lead to some grid points appearing misaligned (Figure \ref{meshcompare}).

\subsection{Performance of the combined model for temperature field prediction}\label{sec42}
In this work, the Gaussian process model has been adopted in the data-driven model for mapping geometric parameters to POD coefficients. Here, the kernel of the Gaussian process is the rational quadratic kernel. To describe the geometric parameters in more detail, as shown in Figure \ref{igd}, curve $P_1P_4$ and $P_5P_6$ are symmetrical about the $x$-axis. Further, make $x_1=2x_2=-2x_3=-x_4$, so only five coordinates $\left(x_1,y_1,y_2,y_3,y_4\right)$ are required to represent the irregular geometric boundary. The five coordinates are considered as the input parameters, and the parameter variation range is shown in Table \ref{para},

\begin{table}[htbp]
	\centering
	\caption{The value range of the five design parameters}
	\begin{tabular}{ccc}
		\toprule
		Parameter & Lower Bound/$mm$ & Upper Bound/$mm$ \\
		\midrule
		$x_1$ & 100 & 150  \\
		$y_1$ & 10 & 16  \\
		$y_2$ & 0 & 30  \\
		$y_3$ & 20 & 50  \\
		$y_4$ & 25 & 75  \\
		\bottomrule
	\end{tabular}
	\label{para}
\end{table}

To demonstrate the effectiveness of the data-driven framework, firstly, the experimental results could explain why using multi-level method rather than single-level, and then compare our model with CNN models. It is worth noting that the temperature field predicted by the data-driven model is a regular area temperature field. Still, to better show the results, the temperature maps we present in this part are all irregular temperatures field, which was created by merging the irregular grid with the regular temperature field.

\subsubsection{On the prediction performance}\label{sec421}
\begin{figure*}[htb]
	\vspace{-0.5cm}
	\setlength{\abovecaptionskip}{-0.3cm}
	\centering
	{\includegraphics[scale=0.7]{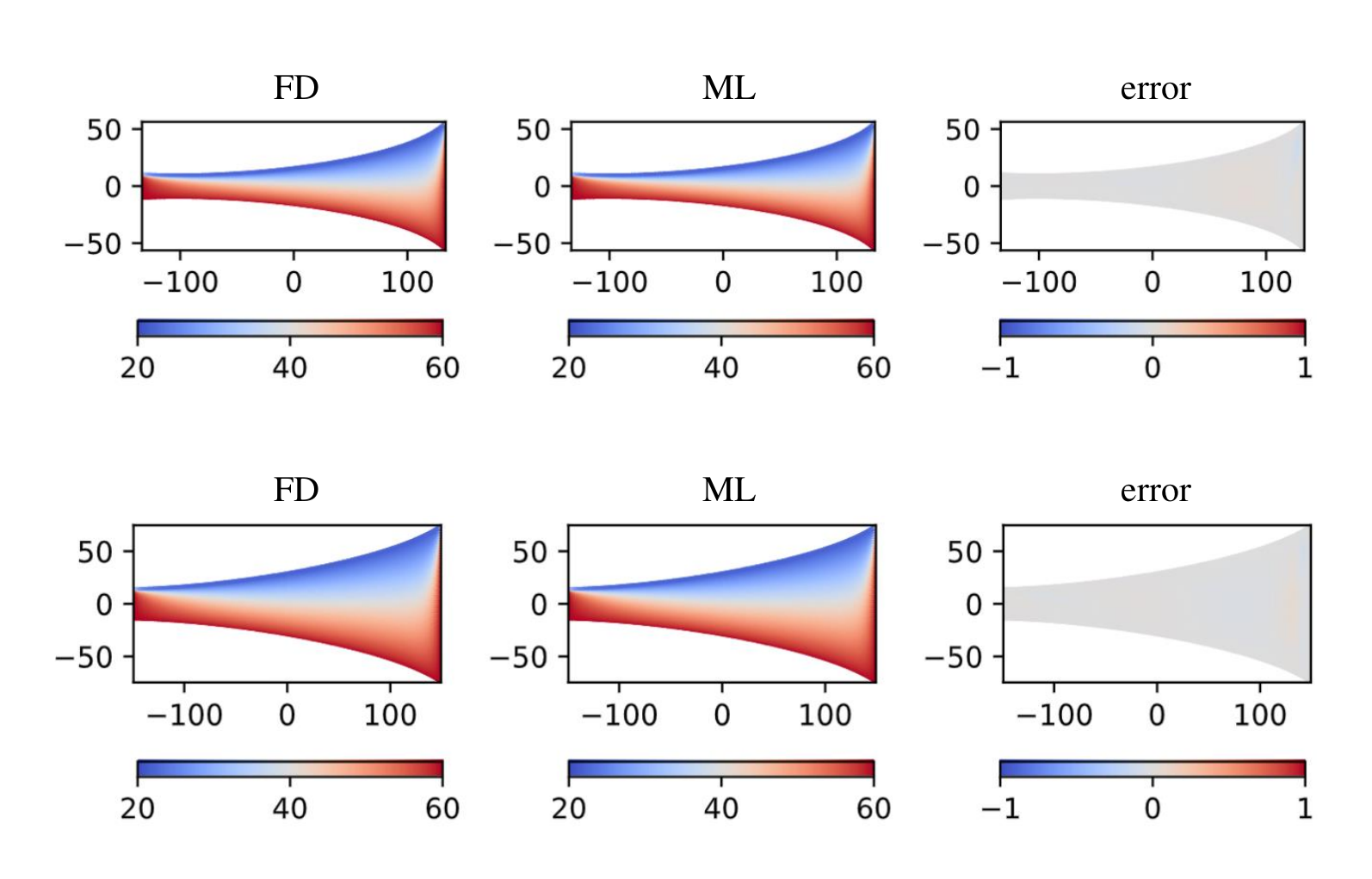}}
	\caption{The prediction performance of two samples in the test set. The "FD" represents the ground-truth temperature field obtained by the finite difference method, the temperature field obtained by our approach is expressed by "ML", and "error" means the error of our method.}\label{mlfd}
\end{figure*}

We first train the multi-level data-driven model on the training set, denoted by $ML$, and then evaluate the model on the test set. $ML$ maintains the MAE of 0.0122K with the MRE of $0.0299\%$ on the test set, and Figure \ref{mlfd} shows the results of two examples in the test set.

\begin{figure*}[htb]
	\vspace{-0.5cm}
	\setlength{\abovecaptionskip}{-0.3cm}
	\centering
	{\includegraphics[scale=0.7]{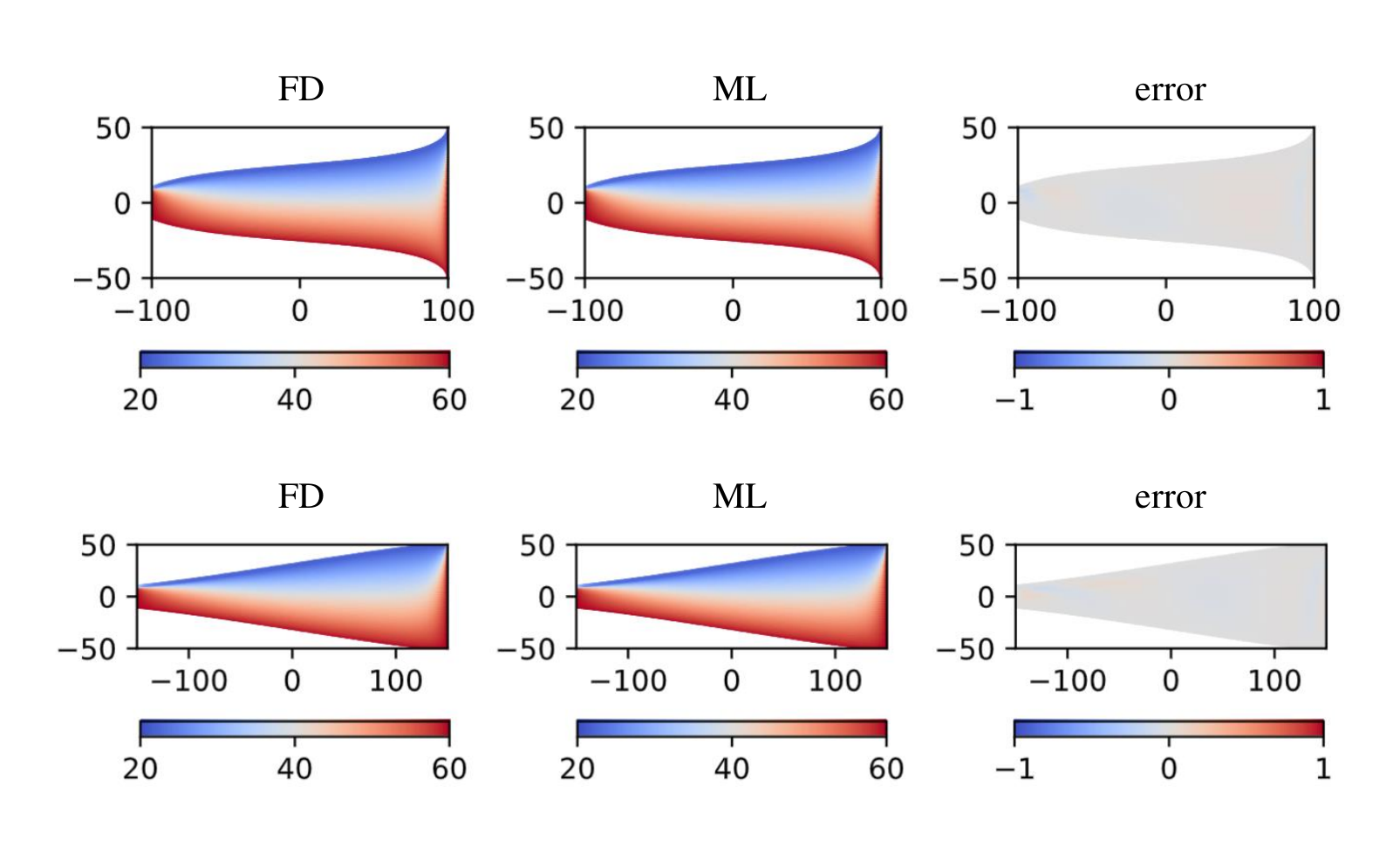}}
	\caption{The prediction performance of two samples on geometric parameters outside the test set range. The "FD" represents the ground-truth temperature field obtained by the finite difference method, the temperature field obtained by our approach is expressed by "ML", and "error" means the error of our method.}\label{mlfd-unseen}
\end{figure*}

It can be seen from Figure \ref{mlfd}, although the geometric boundary changes, our method can still make accurate predictions. We make predictions on geometric parameters outside the test set range either. The results can be seen in Figure \ref{mlfd-unseen}. According to the findings, $ML$ performed well in the test dataset and has a strong generalization performance.

\begin{table}[htbp]
	\centering
	\caption{Performances on the test set for different models}
	\begin{tabular}{cccccc}
		\toprule
		Dataset size & 500 & 1000 & 1500 & 2000 & 2500 \\
		\midrule
		MAE/K  & 0.0157 & 0.0131 & 0.0127 & 0.0123 & 0.0125 \\
		MRE/$\%$ & 0.0378 & 0.0320 & 0.0309 & 0.0299 & 0.0305  \\
		\bottomrule
	\end{tabular}
	\label{scale}
\end{table}

To illustrate the effect of the training data scale on prediction accuracy, we take 500, 1000, 1500, 2500 samples as dataset to train separate Gaussian process models. The performance of the above surrogates is concluded in Table \ref{scale}. According to the results, the scale of the dataset has little effect on prediction accuracy, especially when the number of samples is greater than 1000. In other words, our model can still have high prediction accuracy under a small-scale dataset.

\subsubsection{Comparison with other methods}\label{sec422}

As a novel data-driven model for temperature field prediction, there are two major improvements. One is the utilization of multi-level grids, which enhances the prediction accuracy. The other uses the POD model, translating the high-dimensional temperature field prediction problem into a low-dimensional POD coefficient prediction, which lowers the training complexity. To prove the above two points, we first compare the results with POD usage only on the finest mesh and then compare our model with the CNN models, such as U-net and FNO.

\begin{table}[htbp]
	\centering
	\caption{Performances on the test set for the single-level and multi-level model}
	\begin{tabular}{ccc}
		\toprule
		Surrogate & MAE/K & MRE/$\%$ \\
		\midrule
		$ML$ & 0.0123 & 0.0299  \\
		$SL$ & 0.0185 & 0.0448  \\
		\bottomrule
	\end{tabular}
	\label{mlsl}
\end{table}

\begin{figure*}[b]
	\centering
	{\includegraphics[scale=0.7]{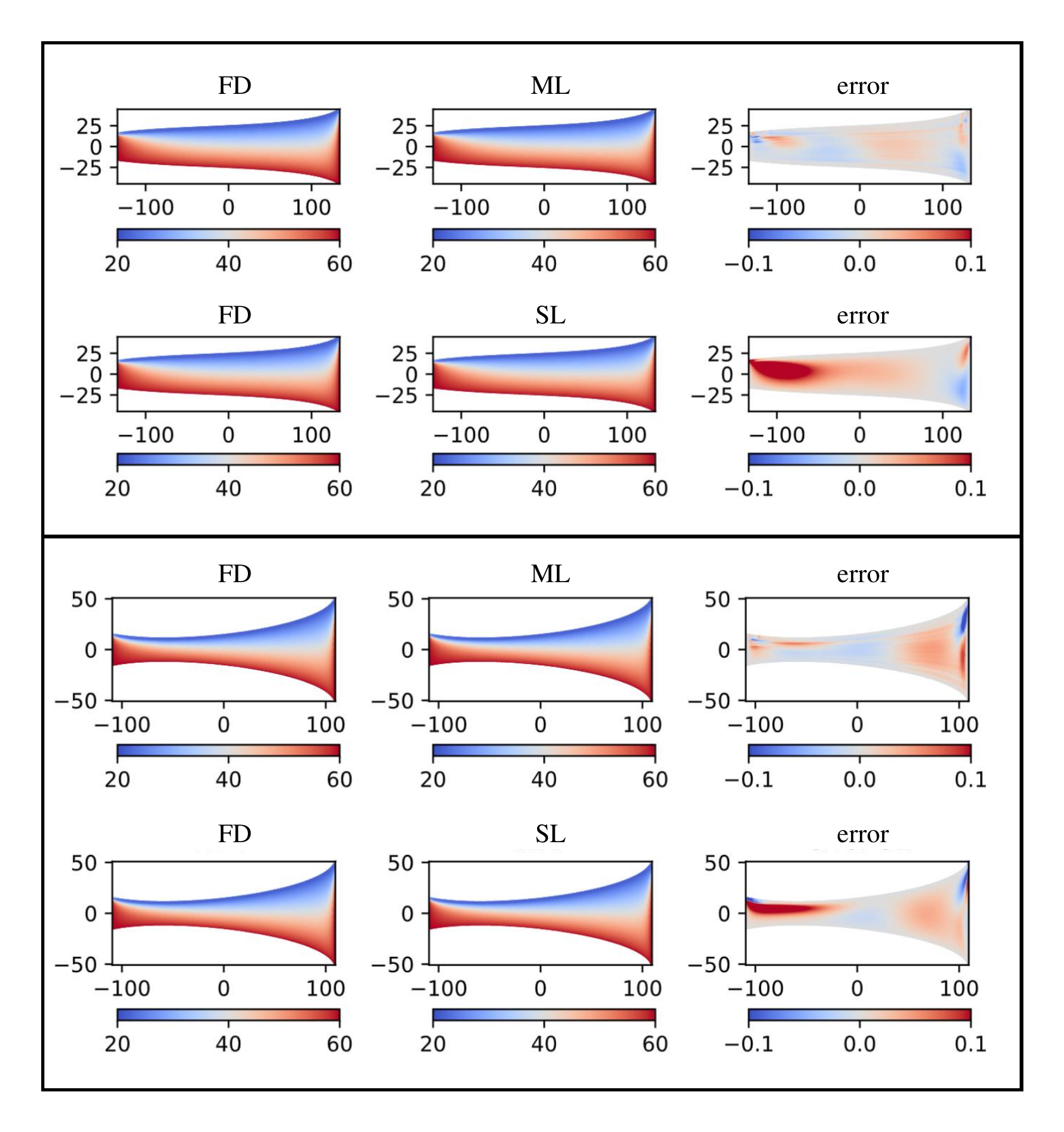}}
	\caption{The prediction performance of single-level and multi-level model on two test samples. The "SL" represents the result predicted by the single-level model, and the temperature field obtained by our approach is expressed by "ML".}\label{slml}
\end{figure*}

\paragraph{\textbf{Comparison with the single finest mesh}}
We investigate the prediction performance of the two models on the same data set and the single-level model denoted by $SL$. The comparisons of MAE and MRE are shown in Table \ref{mlsl} and Figure \ref{slml} show the two samples of predicting results on the test set.

\begin{figure*}[!t]
	\centering
	{\includegraphics[scale=0.7]{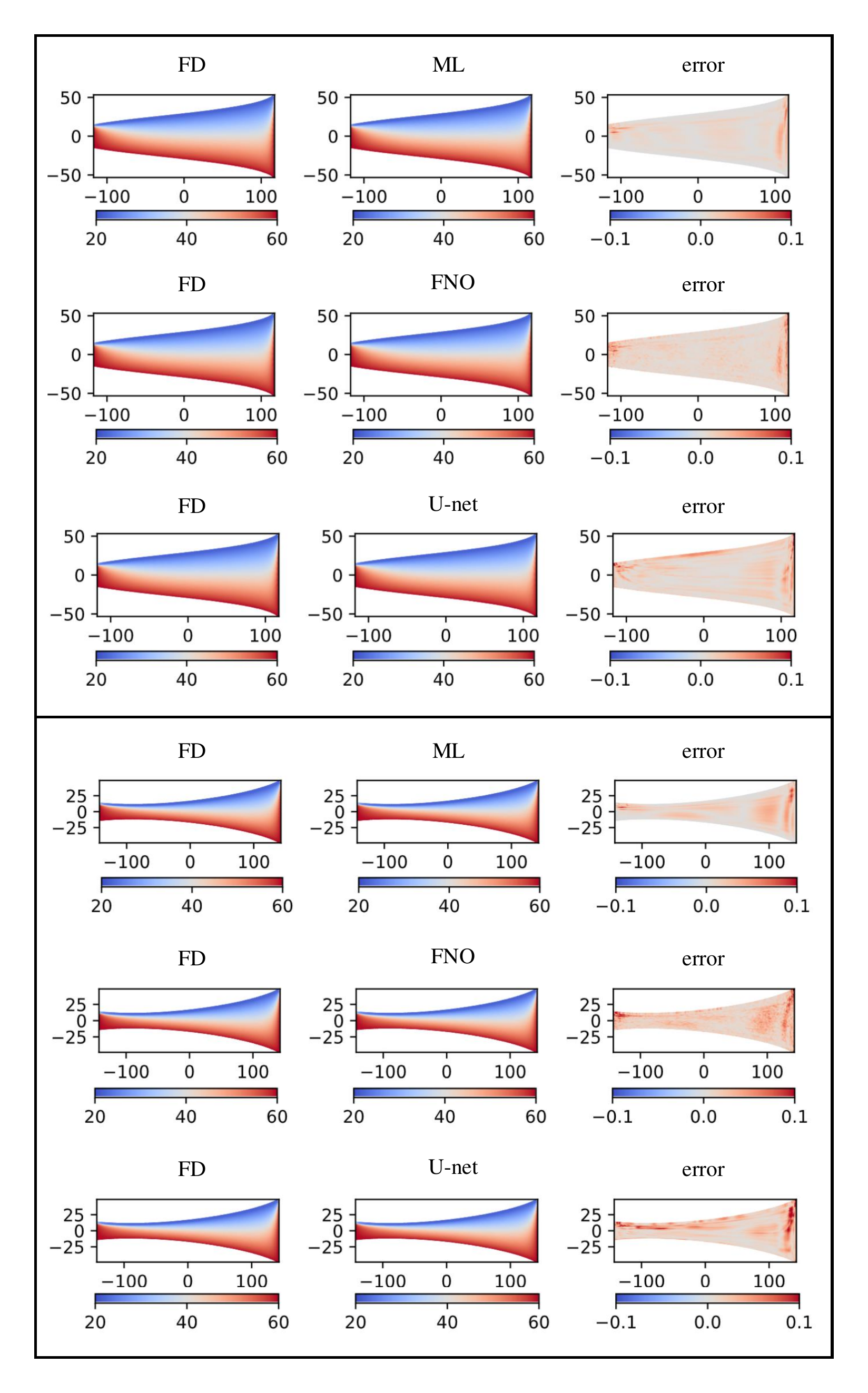}}
	\caption{The prediction performance of three models on two test samples. The temperature field obtained by our approach is expressed by "ML", the "FNO" means the temperature field predicted by the FNO model and the temperature field predicted by the U-net model expressed as "U-net".}\label{fuml}
\end{figure*}

In Figure \ref{slml}, both the single-level and multi-level models can capture the temperature field information. However, notable discrepancies are observed in the single-level predicted temperature field, especially in the region that is near the boundary.  

\paragraph{\textbf{Comparison with the CNN models}} 
The U-net \cite{zhao2021physics} and FNO \cite{li2020fourier} models are employed as a comparison. Specifically, the boundary of the curve is set as model input, while the output of the model is the temperature field.

We first investigate the prediction performance. As can be seen from the error in Figure \ref{fuml}, three models all can accurately predict the temperature field. However, significant differences are observed in the temperature field predicted by U-net and FNO model.

\begin{figure*}[!t]
	\centering
	\subfigure [MAE]
	{\includegraphics[scale=0.4]{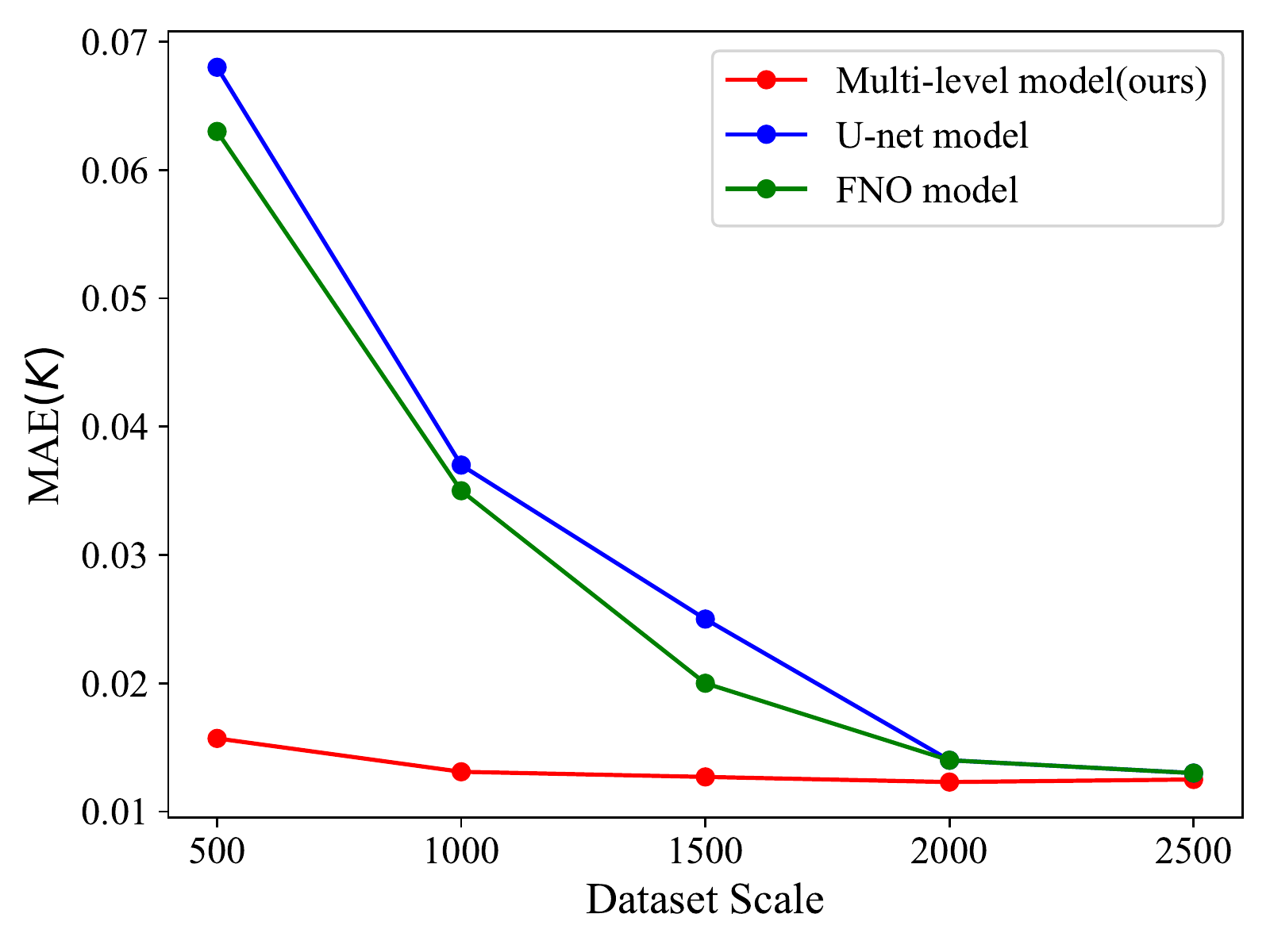}\label{mae}}
	\subfigure [MRE]
	{\includegraphics[scale=0.4]{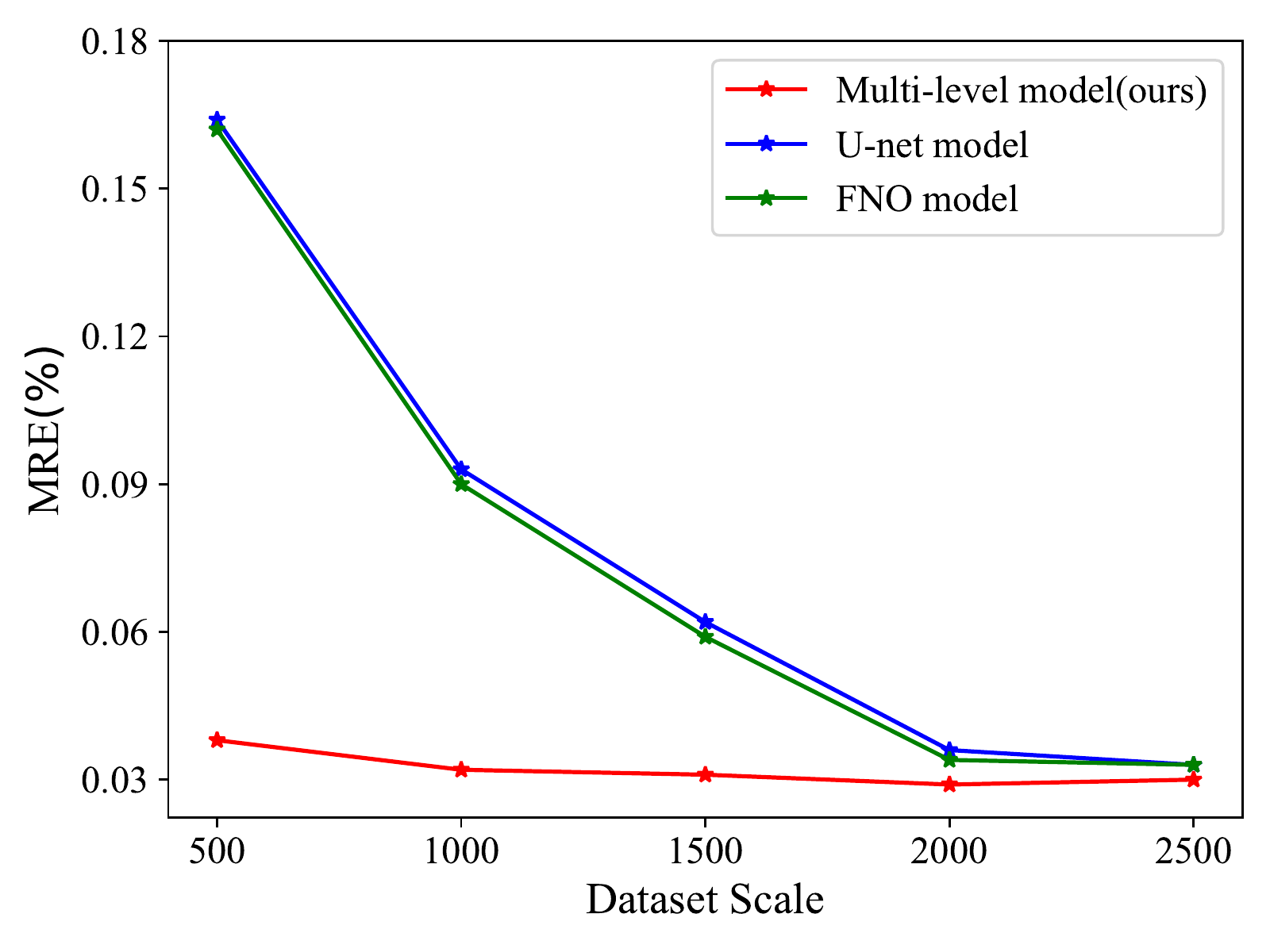}\label{mre}}
	\caption{Performances with different dataset scale for different models.}\label{scaleda}
\end{figure*}

Then, the relationship between the scale of the dataset and the prediction performance is discussed. Without loss of generality, due to the common phenomenon that more training data makes the model less prone to overfitting, we assume that more training data leads to better performance. Figure \ref{scaleda} shows the MAE and MRE of three models with different dataset sizes. It can be observed that the MAE and MRE of the multi-level model are very small equally, and both two indicators have little changes with the increasing of dataset size. In contrast to the U-net and FNO model, the prediction accuracy varies greatly with the scale of the data set, which clearly proves the effectiveness of the proposed multi-level model.

Besides, the training cost has been investigated, observing that the U-net or FNO model needs about 10 times more than the multi-level model does to reach convergence. It takes about 20 minutes to train the U-net or FNO model, but only around 2 minutes to train the multi-level model with higher predictive accuracy.

\section{Conclusions}\label{sec5}
In this paper, a novel physics and data co-driven surrogate model is proposed to deal with the TFP-IGD problem. First, the irregular region is parameterized with the Bezier curve and mapped to the regular computational plane. Then, using a physics-driven meshing surrogate for structure meshing. And a multi-level data-driven thermal surrogate is adopted to predict the temperature field in the mapped regular area.
Finally, we combine the result of the meshing surrogate and thermal surrogate to get the irregular temperature field. Experimental results show that the meshing surrogate can produce meshes of high quality and low error. In addition, the thermal surrogate can make an accurate prediction on smaller datasets. These advantages highlight the potential of the proposed method to solve more complex scalability problems.

Although we explore a novel model for the TFP-IGD problem, many topics still exist for further investigations. Due to the strong generality of our model, it is promising to apply the proposed model to more complex boundary conditions and more intricate problems, even to handle the three-dimensional temperature field prediction problem. Furthermore, it is worth predicting more complex meshes using deep learning models.




%
%

\bibliographystyle{spbasic}
\bibliography{references}

\end{document}